\documentclass[10pt,twocolumn,letterpaper]{article}

\usepackage{wacv}
\usepackage{cite}
\usepackage{times}
\usepackage{epsfig}
\usepackage{graphicx}
\usepackage{algorithmic}
\usepackage{amsmath,amssymb,amsfonts}
\usepackage{enumitem}
\usepackage{url}
\usepackage{booktabs}
\usepackage{multirow}
\usepackage{ifthen}
\usepackage{subcaption}
\usepackage{ulem}
\usepackage{makecell}
\usepackage[utf8]{inputenc}
\usepackage{xcolor,colortbl}
\usepackage{caption}
\usepackage{subcaption}

\newboolean{ack}
\newboolean{combined}
\setboolean{ack}{false}
\setboolean{combined}{true}

\DeclareMathOperator*{\argmax}{argmax} 

\newcommand{\centered}[1]{\begin{tabular}{c} #1 \end{tabular}}
\newcommand{\ft}{\cellcolor[HTML]{6d9eeb}}
\newcommand{\sd}{\cellcolor[HTML]{a4c2f4}}
\newcommand{\td}{\cellcolor[HTML]{c9daf8}}

\usepackage[pagebackref=true,breaklinks=true,letterpaper=true,colorlinks,bookmarks=false]{hyperref}

\wacvfinalcopy  \setboolean{ack}{true}

\ifwacvfinal\fi
\begin{document}

\title{Are Out-of-Distribution Detection Methods Effective on Large-Scale Datasets?}

\author{Ryne Roady$^{1}$ \quad Tyler L. Hayes$^1$ \quad Ronald Kemker$^{1}$ \quad Ayesha Gonzales$^2$ \quad  Christopher Kanan$^{1,3,4}$\\
$^1$Rochester Institute of Tech.\quad $^2$Case Western Reserve University \quad $^3$Paige \quad $^4$Cornell Tech\\
}

\maketitle
\ifwacvfinal\thispagestyle{empty}\fi

\begin{abstract}
Supervised classification methods often assume the train and test data distributions are the same and that all classes in the test set are present in the training set.  However, deployed classifiers often require the ability to recognize inputs from outside the training set as unknowns. This problem has been studied under multiple paradigms including out-of-distribution detection and open set recognition. For convolutional neural networks, there have been two major approaches: 1) inference methods to separate knowns from unknowns and 2) feature space regularization strategies to improve model robustness to outlier inputs.  There has been little effort to explore the relationship between the two approaches and directly compare performance on anything other than small-scale datasets that have at most 100 categories. Using ImageNet-1K and Places-434, we identify novel combinations of regularization and specialized inference methods that perform best across multiple outlier detection problems of increasing difficulty level.  We found that input perturbation and temperature scaling yield the best performance on large scale datasets regardless of the feature space regularization strategy.  Improving the feature space by regularizing against a background class can be helpful if an appropriate background class can be found, but this is impractical for large scale image classification datasets.
\end{abstract}

\section{Introduction}
\begin{figure}[t!]
 \centering
    \includegraphics[width=0.4\textwidth]{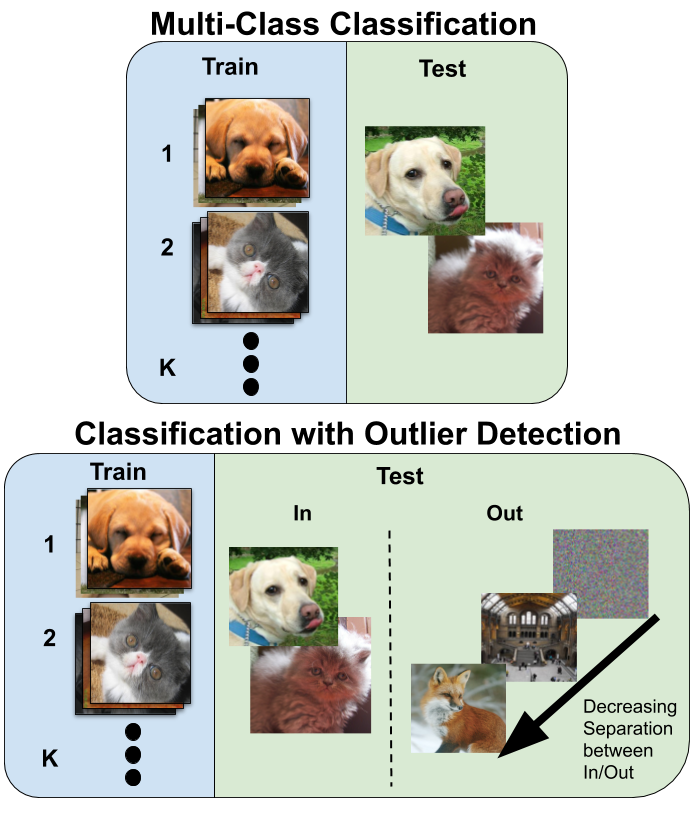}
    \caption{OOD detection is an extension of closed-set classification, where a classifier must determine whether an input is not part of the classes observed during training.  Large-scale image datasets present a difficult challenge for these methods due to the large number of classes and the similarity between the training set distribution and potential unknown classes during inference.  Here, we assess a wide range of OOD detection methods for CNNs on ImageNet-1K and Places-434.
    \label{fig:main-fig}}
    \vspace{-1.5em}
\end{figure}

Convolutional neural networks (CNNs) work extremely well for many categorization tasks in computer vision involving high-resolution images~\cite{he2016deep,hu2018senet}. However, current benchmarks use closed datasets in which the train and test sets have the same classes. This is unrealistic for many real-world applications. It is impossible to account for every eventuality that a deployed classifier may observe, and eventually, it will encounter inputs that it has not been trained to recognize. Out-of-distribution (OOD) detection is the ability for a classifier to reject a novel input rather than assigning it an incorrect label (see Fig.~\ref{fig:main-fig}). This capability is particularly important for the development of 1) safety-critical software systems (e.g., medical applications, self-driving cars) and 2) lifelong learning agents that must automatically identify novel classes to be learned by the classifier~\cite{kemker2018fearnet,parisi2019continual,hayes2019memory,hayes2019remind}.  

For OOD detection in large scale datasets, the major challenge is the presence of `unknown unknowns' since the set of possible distributions of inputs outside of the training set is unbounded.  This problem has been studied under different names including selective classification~\cite{chow1957optimum,el2010foundations}, classification with a reject option~\cite{bartlett2008classification,herbei2006classification}, OOD detection~\cite{hendrycks2017baseline,liang2018enhancing,devries2018learning,lee2018simple}, and open set recognition~\cite{Scheirer2013Towards}. In each case, the goal is to correctly classify inputs that belong to the same distribution as the training set and to reject inputs that are outside of this distribution.  The differences between these names mostly indicate a degree of difference between the training set distribution and the evaluation set containing outlier samples.  In selective classification (or classification with a reject option), the test distribution has the same categories as the training distribution. However, a classifier rejects inputs it cannot confidently classify. In OOD detection, the outlier data used in test cases often comes from entirely different datasets.  In open set recognition, a model is often tested on classifying known classes and reject inputs from classes not observed during training but from the same dataset.  This task requires not only characterizing the input space of the overall training distribution, but often also characterizing the input distribution on a per-class basis.  Surprisingly, there has been little work linking  methods used for each of these paradigms. Here, we consider all of these paradigms to be the same problem with varying levels of difficulty. 

Strategies for OOD detection fall into two general approaches. The first is specialized inference mechanisms for determining if the input to a pre-trained CNN is OOD. The second is to alter the CNN during learning so that it acquires more robust representations of known classes that reduce the probability of a sample from an unknown class being confused.  This often takes the form of collapsing class conditional features in the deep feature space of CNNs.

Finally, the vast majority of prior work for OOD detection in image classification has focused on small, low-resolution datasets, e.g., MNIST and CIFAR-100. Deployed systems like autonomous vehicles, where outlier detection would be critical, often operate on images that have far greater resolution and experience environments with far more categories. It is not clear from previous work if existing methods will scale.  In this paper we compare methods across bounded classification paradigms on large-scale, high-resolution image datasets.

\textbf{Our major contributions are:}
\begin{itemize}[noitemsep,nolistsep]
 \item We organize OOD detection methods for CNNs into families with a shared framework.
 \item We are the first to directly compare inference methods and feature space regularization strategies for OOD and open-set recognition to quantify the benefit gained from combining these techniques.
 \item We extensively compare combinations of inference and feature space methods, many of which have not been previously explored. 
 \item Using ImageNet-1K and Places-434, we find that the performance benefit from feature space regularization strategies decreases for large-scale datasets evaluated in an open set recognition paradigm. At this time these techniques which add complexity to training do not considerably outperform the baseline feature space acquired through standard cross-entropy training. 
\end{itemize}

\section{Problem Formulation}
While OOD detection is related to uncertainty estimation~\cite{hendrycks2018benchmarking} and model calibration~\cite{guo2017calibration}, its function is to reject inappropriate inputs to the CNN.  We formulate the problem as a variant of traditional multi-class classification where an input belongs to either one of the $K$ categories from the training data distribution or to an outlier/rejection category, which is denoted as the $K+1$ category. Given a training set $D_{train} = \left\{ {\left( {{{X}}_1 ,y_1 } \right),\left( {{{X}}_2 ,y_2 } \right), \ldots ,\left( {{{X}}_n ,y_n } \right)} \right\}$, where ${X}_i$ is the $i$-th training input tensor and $y_i \in C_{train} = \left\{1, 2, \ldots, K  \right\} $ is its corresponding class label, the goal is to learn a classifier $F\left(X\right) = (f_1,...,f_k)$, that correctly identifies the label of a known class and separates known from unknown examples: 
 \begin{equation}
 \label{eq:bounded-classification-formulation}
     \hat{y} = \left\{\begin{array}{ll}\argmax_{k} F(X) & \mbox{if} {S}({X}) \geq \delta \\ K+1 & \mbox{if }{S}({X}) < \delta \end{array} \right.
 \end{equation}
where ${S} \left({X} \right)$ is an acceptance score function that determines whether the input belongs to the training data distribution and $\delta$ is a threshold.  

For testing, the evaluation set contains samples from both the set of classes seen during training and additional unseen classes, i.e., $D_{test} = \left\{ {\left( {{{X}}_1 ,y_1 } \right),\left( {{{X}}_2 ,y_2 } \right), \ldots ,\left( {{{X}}_n ,y_n } \right)} \right\}$, where $y_i \in  \left( C_{train}  \bigcup C_{unk} \right)$ and $C_{unk}$ contains classes that are not observed during training. 

\section{OOD Detection in CNNs}
\label{sec:alg-review}
We have organized methods for OOD detection into two complementary families: 1) inference methods that create an explicit acceptance score function for separating outlier inputs, and 2) regularization methods that alter the feature representations during training to better separate in-distribution and OOD samples.  

\subsection{Inference Methods}
Inference methods use a pre-trained neural network to perform OOD detection, but modify how the network outputs are used. Using pre-trained networks is advantageous since no modifications to training need to be made to handle outlier samples, and the low-level features of pre-trained networks have been shown to generalize across different image datasets~\cite{yosinski2014transferable}. 

\subsubsection{Output Layer Thresholding}
The simplest approach to OOD detection is thresholding the output of a model, typically after normalizing by a softmax activation function. For multi-class classifiers, the softmax layer assumes mutually exclusive categories, and in an ideal scenario would produce a uniform posterior prediction for a novel sample. Unfortunately, this ideal scenario does not occur in practice and serves as a poor estimate for uncertainty~\cite{nguyen2015deep,gal2016dropout}. Still, the largest output of the softmax layer follows a different distribution for OOD examples, i.e., in-distribution samples generally have a much larger top output than OOD samples, and can be used to reject them~\cite{hendrycks2017baseline}. We refer to this output thresholding method as $\tau$-Softmax.

The Out-of-Distribution Image Detection in Neural Networks (ODIN) model~\cite{liang2018enhancing} extends the thresholding approach by adjusting the softmax output through temperature scaling on the activation function. ODIN also applies small input perturbations to the test samples based on the gradient of this temperature adjusted softmax output. In this application, the sign of the gradient is used to enhance the probability of inputs that are in-distribution while minimally adjusting the output of OOD samples.

Additionally, per-class thresholds can be set for sample rejection typically after using a sigmoid activation function on the output logit.  The sigmoid activation helps to avoid the normalization properties of the softmax activation and create more discriminative per-class thresholds. This method is employed in the Deep Open Classification (DOC) model~\cite{shu2017deep}, which alters a typical multi-class CNN architecture by replacing the softmax activation of the final layer with a one-vs-rest layer containing $K$ sigmoid functions for the $K$ classes seen during training.  A threshold, $k_i$, is then established for each class by treating each example where $y=k_i$ as a positive example and all samples where $y\neq k_i$ as negative examples.  During inference, if all outputs from the sigmoid activations are less than the respective per-class thresholds, then the sample is rejected.  For our evaluations, we separate this per-class thresholding strategy from the one-vs-rest model training strategy to isolate the benefits of each method. 

\subsubsection{Distance Metrics}
Outlier detection can also be done using distance-based metrics. Following the formulation of Knorr and Ng~\cite{knox1998algorithms}, a number of distance-based methods~\cite{aggarwal2001outlier, bay2003mining, angiulli2002fast, qin2010vod} have been developed based on global and local density estimation by
computing the distance between a sample and the underlying data distribution.  

Euclidean distance metrics have been widely used~\cite{taigman2014deepface, schroff2015facenet}, but they often fail in high-dimensional feature spaces containing many classes. To mitigate this issue, \cite{murphy2012machine} showed that the feature space of a neural network trained with cross-entropy loss approximates a Gaussian discriminant analysis classifier with a tied covariance matrix between classes.  Under this assumption, a Mahalanobis distance metric can be used for generating a class-conditional outlier score from the deep features in a CNN.

This approach is employed directly on CNNs by the Mahalanobis method~\cite{lee2018simple}, which computes a class-conditional Mahalanobis metric across multiple CNN layers and learns a linear classifier to combine these into a single acceptance score based on cross-fold validation.  

\subsubsection{One-Class Networks}
Another technique for learning a decision boundary in feature space to separate in-distribution data from outlier data is a one-class classifier.  The most popular one-class techniques are currently based on Support Vector Machines (SVM)~\cite{scholkopf2001estimating, scholkopf2000support}, with recent work focused on learning features that enable anomaly detection~\cite{erfani2016high, perera2019learning}.  One-class SVMs find the maximum margin decision boundary such that some portion of training samples fall inside the boundary.  The estimate of the proportion of training data that should be considered as the `outlier' class is a hyper-parameter that must be set through cross-validation.
    
\subsubsection{Extreme Value Theory}
OOD detection methods based on extreme value theory (EVT) recognize novel inputs by characterizing the probability of occurrences that are more extreme than any previously observed.  This is typically implemented by characterizing the tail of class-conditional distributions in feature space.  It has been directly adapted to CNN classifiers by modeling the distance to the nearest class mean in deep feature space as an extreme value distribution~\cite{scheirer2011meta, scheirer2014probability} and calculating an acceptance score function as the posterior probability based on this EVT distribution. OpenMax~\cite{bendale2016opensetdeep} specifically applies EVT to construct a sample weighting function to re-adjust the output activations of a CNN based on a per-class Weibull probability distribution.  The output is rebalanced between the closed set classes and a rejection class, and samples are rejected if the rejection class has a maximum activation or if the maximum activation falls below a threshold set from cross-fold validation.

\begin{table}[b]
\centering
\caption{The studied inference methods for OOD detection.  Inference complexity refers to the number of passes through a deep CNN (forward and backward) during inference. }
\resizebox{0.5\textwidth}{!}{
\begin{tabular}{lcc}
\toprule
\thead{\textsc{Classification} \\ \textsc{Method}} & \thead{\textsc{Acceptance Score} \\ \textsc{Function}} & \thead{\textsc{Inference} \\ \textsc{Complexity}} \\ \midrule
$\tau$-Softmax~\cite{hendrycks2017baseline} & Simple Threshold & 1 \\
DOC~\cite{shu2017deep} & Per-Class Threshold & 1\\
ODIN~\cite{liang2018enhancing} & Temp Adjusted Threshold & 3\\
OpenMax~\cite{bendale2016opensetdeep} & Per-Class EVT Rescaling & 1 \\
One-Class SVM~\cite{scholkopf2001estimating} & SVM Score & 1 \\
Mahalanobis~\cite{lee2018simple} & Generative-Distance Metric & 3 \\
\bottomrule
\end{tabular} %
}
\label{table:model_summary}
\end{table}

\subsection{Feature Representation Methods}

In contrast to methods that solely use the acceptance score function, feature representation methods alter the architecture of the network or how the network is trained. These methods learn representations that enable better OOD detection performance.

\subsubsection{One-vs-Rest Classifiers}
The most common method for training a CNN classifier with $K$ disjoint categories is using cross-entropy loss calculated from a softmax activation function.  Although the softmax function is good for training a classifier over a closed set of classes, it is problematic for outlier detection because the output probabilities are normalized, resulting in high-probability estimates for inputs that are either absurd or intentionally produced to fool a network~\cite{goodfellow2014explaining, nguyen2015deep}.  One-vs-rest classification models eliminate the softmax layer of a traditional closed-set classifier and replace it with a logistic sigmoid function for each class. While these per-class sigmoid activations no longer have a probabilistic interpretation in a multi-class problem, they reduce the risk of incorrectly classifying an OOD sample by treating each class as a closed-set classification task, which can be individually thresholded to identify outliers.  The DOC model is one version of a one-vs-rest classifier that replaces the traditional softmax layer with a one-vs-rest layer of individual logistic sigmoid units~\cite{shu2017deep}.

\subsubsection{Background Class Regularization} \label{sec:bkg_class}
Another method for improving OOD detection performance via feature space regularization is using a background class to separate novel classes from known training samples.  This technique is most commonly applied in object detection algorithms where the use of separate region proposal and image classification algorithms result in a classifier that must handle ambiguous object proposals~\cite{ren2015faster}.  Often these classifiers represent the background class as a separate output node which is trained using datasets that have an explicit `clutter' class such as MS COCO~\cite{lin2014microsoft} or Caltech-256~\cite{griffin2007caltech}.  Alternatively, newer approaches have used background samples to train a classifier to predict a uniform distribution when presented with anything other than an in-distribution training sample~\cite{hendrycks2018deep}.  This is done through various regularization schemes including confidence loss~\cite{lee2017training} and the objectosphere loss~\cite{dhamija2018reducing} which have shown better performance than using a separate output node.  Nevertheless, for modern image classification datasets which may have 1,000+ classes, finding explicit background samples that are exclusive of the training classes has become exceedingly difficult.

\subsubsection{Generative Models} \label{sec:generative}
Using CNNs for generative modeling has been an active area of research with the advent of generative adversarial networks~\cite{goodfellow2014generative} and variational auto-encoders~\cite{kingma2013auto}.  Generative models have extended earlier density estimation approaches for outlier detection by more accurately approximating the input distribution. A well-trained model can be used to directly predict if test samples are from the same input distribution~\cite{nalisnick2018deep} or estimate this by measuring reconstruction error~\cite{oza2019deep}.  Paradoxically, generative models have also been used to create OOD inputs from the training set in order to condition a classifier to produce low confidence estimates similar to how an explicit background class is used for model regularization~\cite{yu2017open, lee2017training, neal2018open}.  

\section{Methods Assessed}
We compare six of the inference methods described in Sec.~\ref{sec:alg-review} on large-scale image classification datasets trained using one of three different feature space regularization strategies. We chose these methods and strategies based on their ability to scale to large datasets. For this reason, we omitted some recent generative methods (e.g., \cite{yu2017open, lee2017training, neal2018open}) because of convergence difficulty and instability during training~\cite{goodfellow2014generative} on large-scale datasets. We also omit the ensemble of multiple leave-out classifiers method~\cite{vyas2018out}, which requires training 1,000 and 434 separate classifiers for the ImageNet and Places datasets respectively.

\begin{figure*}[t!]
\resizebox{\textwidth}{!}{
\setlength\tabcolsep{0.1mm}
\begin{tabular}{ccccccc}
      & \centered{Test Data and\\Class Boundaries} & \centered{Baseline\\Thresholding} & \centered{Temperature\\Scaling} & \centered{EVT on Output\\(OpenMax)} & \centered{One-Class\\SVM} & \centered{Mahalanobis Dist\\Thresholding} \\
     \centered{Cross-Entropy} & \centered{\includegraphics{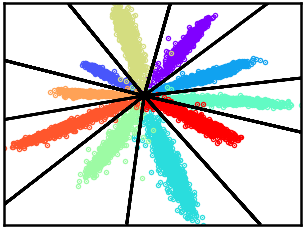}} & \centered{\includegraphics{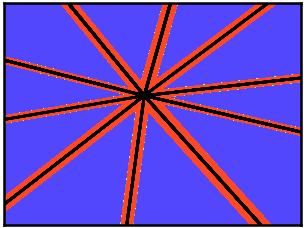}} & \centered{\includegraphics{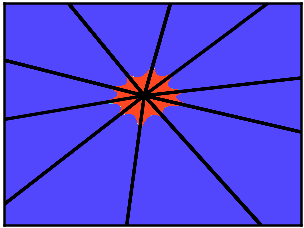}} & \centered{\includegraphics{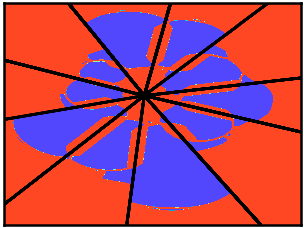}} &     \centered{\includegraphics{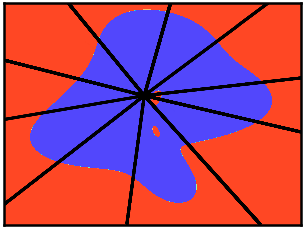}} &     \centered{\includegraphics{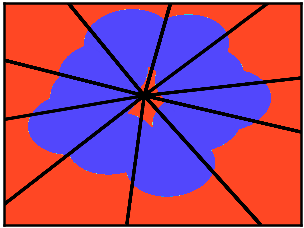}} \\
    \centered{One-vs-Rest \\(Binary Cross-Entropy)} &  \centered{\includegraphics{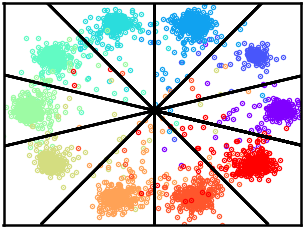}} & \centered{\includegraphics{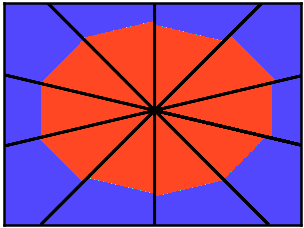}} & \centered{\includegraphics{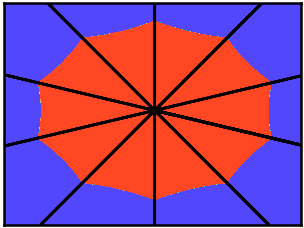}} & \centered{\includegraphics{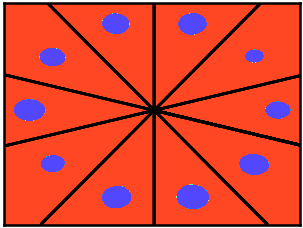}} &     \centered{\includegraphics{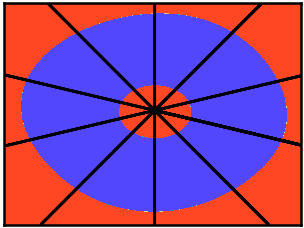}} &    \centered{\includegraphics{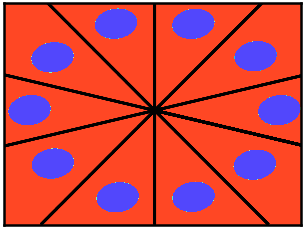}}  \\
    \centered{Background Class\\Regularization} & \centered{\includegraphics{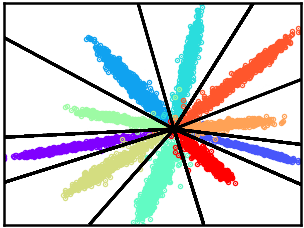}} & \centered{\includegraphics{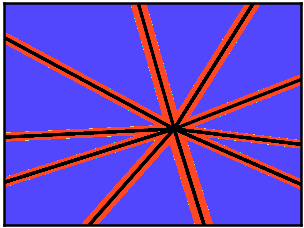}} & \centered{\includegraphics{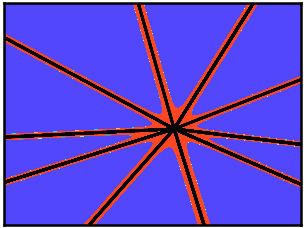}} & \centered{\includegraphics{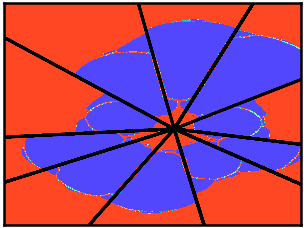}} &     \centered{\includegraphics{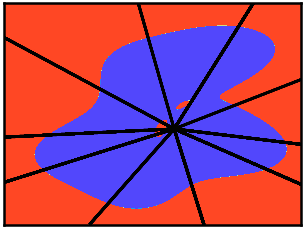}} &
    \centered{\includegraphics{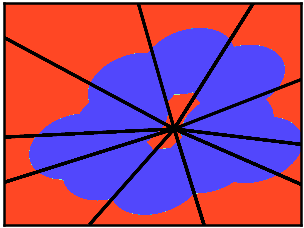}}  \\
\end{tabular}
}
\caption{2-D visualization of the decision boundaries created from the different OOD inference methods studied using the LeNet+ architecture and MNIST~\cite{lecun1998gradient} as the training set.  \textbf{\textcolor[HTML]{0000FF}{Blue}} is the acceptance region for in-distribution samples calibrated at a 95\% True Positive Rate (TPR) for training data.  \textbf{\textcolor[HTML]{ff0000}{Red}} is the rejection region (outlier).}
\label{fig:decision_boundaries}
\end{figure*} 

\begin{figure*}[t]
    \centering
    \begin{subfigure}[b]{0.2\textwidth}
        \centering
        \includegraphics[width=\textwidth]{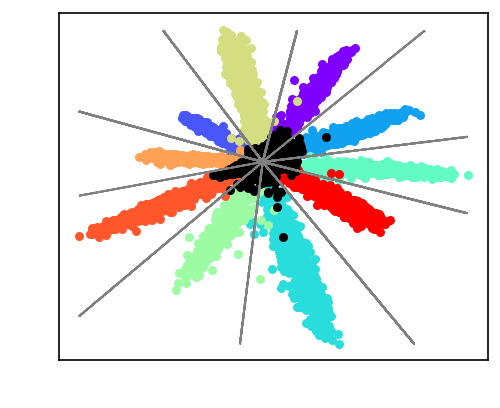}
        \caption{Cross-Entropy}
        \label{fig:ood_ce}
    \end{subfigure}
    \hspace{1mm}
    \begin{subfigure}[b]{0.2\textwidth}
        \centering
        \includegraphics[width=\textwidth]{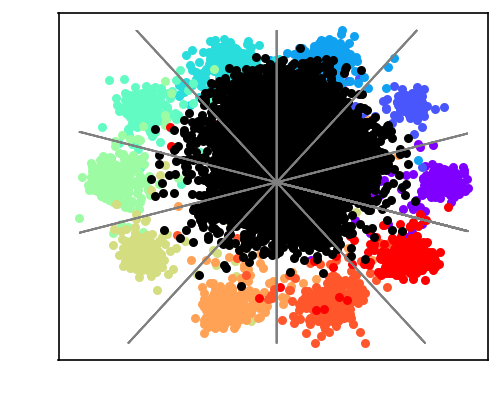}
        \caption{One-vs-Rest}
        \label{fig:ood_bce}
    \end{subfigure}
    \hspace{1mm}
    \begin{subfigure}[b]{0.2\textwidth}
        \centering
        \includegraphics[width=\textwidth]{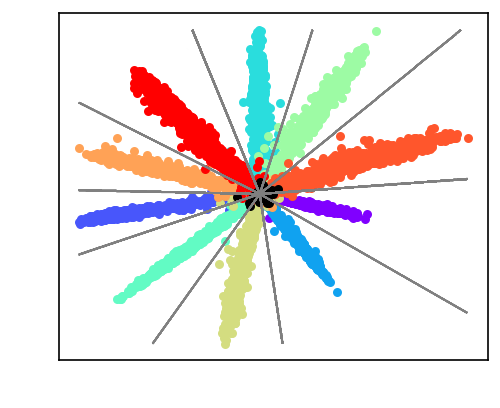}
        \caption{Background Class Reg.}
        \label{fig:ood_bkg}
    \end{subfigure}
    \caption{2-D visualization of the effect of the different feature space regularization strategies on separating in-distribution and outlier inputs The in-distribution training set is \textcolor{red}{M}\textcolor{orange}{N}\textcolor[HTML]{CCCC00}{I}\textcolor[HTML]{008000}{S}\textcolor{blue}{T} while the OOD set is \textbf{Fashion-MNSIT}~\cite{xiao2017fmnist}.  For background class regularization, the EMNIST-Letters dataset~\cite{cohen2017emnist} is used as a source for background samples}
    \label{fig:ood_three_comp}
    \vspace{-1.5em}
\end{figure*}


\subsection{Inference Methods}

Specific implementation details for the inference methods evaluated are as follows:
\begin{enumerate}[noitemsep,nolistsep]
    \item \textbf{$\tau$-Softmax, $\tau$-Sigmoid} -- This simple baseline approach finds a global threshold from the final output of the model after the associated activation function is applied.  The method yields good results on common small-scale datasets~\cite{hendrycks2017baseline} and can be easily extended to datasets with many classes.
    
    \item \textbf{DOC} -- Per-class thresholding has been shown to successfully reject outlier inputs during testing on common, small-scale datasets~\cite{shu2018unseen}.  Adapting this method to larger datasets is more computationally expensive than  $\tau$-Softmax  because a per-class threshold must be established.
    
    \item \textbf{ODIN} -- This approach can outperform $\tau$-Softmax when using well-trained CNNs; however, the technique adds computational complexity during inference to calculate input perturbations~\cite{liang2018enhancing}. ODIN also adds additional hyperparameters for the magnitude of input perturbation and a temperature scaling factor which must be determined through cross validation. 
    
    \item \textbf{OpenMax} -- OpenMax is one of the only methods previously tested on ImageNet-1K~\cite{bendale2016opensetdeep}. It models a per-class EVT distribution and has multiple hyperparameters that must be tuned through cross validation making it relatively cumbersome so use for large-scale datasets during training.  Once these parameters have been found, however, it presents a relatively straight-forward inference method for estimating whether a sample belongs to one of the known classes or to an explicitly modelled outlier class.  

    \item \textbf{One-Class SVM} -- One-class SVMs have been employed as a simple unsupervised alternative to density estimation for detecting anomalies.  They have been tested across a wide variety of datasets, but not on the large-scale image datasets and CNN architectures used in this analysis.  We use a radial basis function kernel to allow a non-linear decision boundary in deep feature space and tune hyperparameters via cross-validation.
    
    \item \textbf{Mahalanobis} -- In \cite{lee2018simple}, the Mahalanobis metric was computed at multiple layers within a network and then combined via a linear classifier that was calibrated using a small validation set made up of in-distribution and OOD samples. To avoid biasing the model by training with OOD data, we only compute the Mahalanobis metric in the final feature space. Adapting this metric to a large-scale dataset is straightforward, however, there is additional computational and memory overhead to estimate and store class conditional means and a global covariance matrix in feature space.

\end{enumerate}

We use the $\tau$-Softmax, $\tau$-Sigmoid, ODIN, and OpenMax methods without modification, while Mahalanobis is modified to only compute distance in the final feature space. Hyperparameters for each inference method are tuned using outlier samples drawn from Gaussian noise to avoid unfairly biasing results to the datasets used for evaluation.

\subsection{Feature Space Methods}
The feature space regularization strategies for improving outlier detection were implemented as follows:
\begin{enumerate}[noitemsep,nolistsep]
    \item \textbf{Cross-Entropy} -- As a baseline, we train each network with standard cross-entropy loss to represent a common feature space for CNN-based models. 
    
    \item \textbf{One-vs-Rest} -- The one-vs-rest training strategy was implemented by substituting a sigmoid activation layer for the typical softmax activation and using a binary cross-entropy loss function.  In this paradigm, every image is a negative example for every category it is not assigned to. This creates a much larger number of negative training examples for each class than positive examples.  For this reason, we re-weight the negative-class training loss to be proportional to the positive-class loss to ensure comparable closed-set validation accuracy.

    \item \textbf{Background Class Regularization} -- The Entropic Open Set method~\cite{dhamija2018reducing} is a regularization scheme which uses a background class and a unique loss function during training to optimize the feature space of a neural network for separating known classes from potential unknowns.  The entropic open set loss forces samples from the background class to the null vector in feature space by calculating the cross-entropy of a uniform distribution for these samples similarly to the confidence loss term in~\cite{lee2017training}.  An additional regularization term is used to measure the hinge loss of the magnitude between samples in the background class and the training samples in feature space.  For ImageNet, we use samples drawn from exclusive classes in the Places dataset as a background class, and vice versa for Places with ImageNet.  Overlapping classes between the two datasets were removed from the training and evaluation datasets.
    
\end{enumerate}

For both ImageNet and Places, we train for 90 epochs starting with a learning rate of 0.1 decayed by a factor of 10 every 30 epochs. Stochastic gradient descent with momentum of $0.9$ and weight decay of $5e-5$ were used.  All training parameters were held constant for all feature space regularization strategies unless otherwise noted.

\section{Qualitative Analysis: Feature Space Visualization}

To visually illustrate the differences between various methods, we trained a simple model for outlier detection using the MNIST dataset. We used a shallow CNN with a bottle-necked feature layer, i.e., the LeNet++ architecture~\cite{lecun1998gradient}, to allow visualization of the resulting decision boundaries. Fig.~\ref{fig:decision_boundaries} shows the 2-D decision boundaries with blue representing in-distribution classification at a 95\% true positive rate threshold and red representing the resulting rejection region.  Additionally, we mapped samples from an unknown class represented by the Fashion-MNIST~\cite{xiao2017fmnist} dataset in Fig.~\ref{fig:ood_three_comp} to understand how the decision boundaries relate to the deep CNN features of known and unknown classes.

These results illustrate that for a given feature space, inference strategies can be divided between those that have unbounded acceptance regions (e.g., $\tau$-Softmax) with those that are bounded (e.g., OpenMax). Much has been made of this distinction~\cite{Scheirer2013Towards} and it is seen as a strength of the inference methods with bounded regions. However, as Fig.~\ref{fig:ood_three_comp} represents, unknown inputs are rarely mapped into these unbounded regions, but rather are centered around the origin in the deep feature space of a CNN. This implies that properly mapping the acceptance/rejection region around the origin is critical performance. Of the bounded acceptance region methods, OpenMax and Mahalanobis create the most compact decision boundaries. However, having compact boundaries may not be the best option when generalization to test inputs and unknown novel inputs is desired.

The goal of different feature space regularization strategies is to build robustness into the deep feature space by separating knowns from potential unknowns.  While naively the One-vs-Rest training strategy appears to be a good solution by creating more compact class conditional distributions, the technique does not directly impact how features from unknown inputs will be mapped into the deep feature space.  Instead we see that regularizing the model with a representation of the unknown class creates better separation between the known and unknown~\cite{lee2017training,dhamija2018reducing}.  The difficulty in this approach, however, lies in large-scale datasets with many hundreds of classes.

\begin{table*}[t]
\centering
\caption{AUROC results averaged over 5 runs for the methods tested.  Top performer for each in-distribution / out-of-distribution combination is in \colorbox[HTML]{a4c2f4}{blue} along with statistically insignificant differences from the top performer as determined by DeLong's test~\cite{delong1988comparing} 
($\alpha = 0.01$ with a correction for multiple comparisons within each column).}
\label{tab:results_w_delong}
\resizebox{\textwidth}{!}{%
\begin{tabular}{c|r|ccc|ccc|}
\multicolumn{1}{l|}{} & \multicolumn{1}{l|}{} & \multicolumn{3}{c|}{\textbf{ImageNet}} & \multicolumn{3}{c|}{\textbf{Places}} \\ \hline
\textbf{Features Space} & \multicolumn{1}{l|}{\textbf{Inference Method}} & \textbf{\begin{tabular}[c]{@{}c@{}}Gaussian \\ Noise\end{tabular}} & \textbf{\begin{tabular}[c]{@{}c@{}}Inter-Dataset \\ (OOD)\end{tabular}} & \textbf{\begin{tabular}[c]{@{}c@{}}Intra-Dataset \\ (Open Set)\end{tabular}} & \textbf{\begin{tabular}[c]{@{}c@{}}Gaussian\\ Noise\end{tabular}} & \textbf{\begin{tabular}[c]{@{}c@{}}Inter-Dataset\\ (OOD)\end{tabular}} & \textbf{\begin{tabular}[c]{@{}c@{}}Intra-Dataset\\ (Open Set)\end{tabular}} \\ \hline
\multirow{6}{*}{CrossEntropy}
 & $\tau$-Softmax & 0.976 & 0.823 & 0.785 & 0.758 & 0.604 & 0.589 \\
 & DOC & 0.975 & 0.825 & 0.786 & 0.759 & 0.604 & 0.589 \\
 & ODIN & \sd 1.000 & 0.906 & \sd 0.852 & 0.889 & 0.499 & 0.474 \\
 & OpenMax & 0.855 & 0.792 & 0.741 & 0.992 & 0.797 & 0.625 \\
 & One-Class SVM & 0.985 & 0.828 & 0.696 & 0.804 & 0.624 & 0.617 \\
 & Mahalanobis & 0.886 & 0.592 & 0.689 & 0.996 & 0.693 & \sd 0.714 \\ \hline
\multirow{6}{*}{One vs Rest}
 & $\tau$-Sigmoid & \sd 0.998 & 0.737 & 0.698 & \sd 0.999 & 0.636 & 0.639 \\
 & DOC & 0.951 & 0.665 & 0.648 & \sd 0.999 & 0.635 & 0.637 \\
 & ODIN & \sd 1.000 & 0.815 & 0.740 & \sd 0.999 & 0.627 & 0.633 \\
 & OpenMax & 0.809 & 0.702 & 0.642 & \sd 1.000 & 0.635 & 0.638 \\
 & One-Class SVM & 0.981 & 0.757 & 0.626 & 0.829 & 0.664 & 0.672 \\
 & Mahalanobis & 0.951 & 0.638 & 0.688 & 0.996 & 0.649 & 0.685 \\ \hline
\multirow{6}{*}{Background Class Regularization}
 & $\tau$-Softmax & 0.905 & 0.910 & 0.795 & 0.992 & 0.860 & 0.600 \\
 & DOC & 0.911 & 0.911 & 0.794 & 0.992 & 0.860 & 0.600 \\
 & ODIN & \sd 0.999 & \sd 0.957 & \sd 0.856 & \sd 0.998 & \sd 0.912 & 0.643 \\
 & OpenMax & 0.920 & 0.837 & 0.761 & 0.985 & 0.855 & 0.597 \\
 & One-Class SVM & 0.978 & 0.944 & 0.737 & 0.976 & 0.901 & 0.655 \\
 & Mahalanobis & 0.886 & 0.403 & 0.608 & 0.805 & 0.485 & 0.660 \\ \hline
\end{tabular}%
}
\end{table*}

\section{Empirical Analysis on Large-Scale Datasets} \label{sec:empirical}

\subsection{Datasets \& Evaluation Paradigm}
To estimate the ability of OOD detection methods to scale, we trained models on two large-scale image classification datasets: ImageNet-1K and Places-434. The ImageNet-1K dataset was part of the ILSVRC challenge between 2012 and 2015 and evaluated an algorithm's ability to classify inputs into 1,000 categories.  The dataset consists of 1.28 million training images (732-1300 per class) and 50,000 evaluation images (50 per class).  Places-434 is an extension of the Places-365 dataset with additional data for 69 categories.  The dataset was used as part of the Places Challenge whose goal was to identify scene categories depicted in images.  The dataset consists of 1.90 million training and 43,100 evaluation images.  For each dataset, we train a ResNet-18~\cite{he2016deep} model on half of the dataset's classes, i.e., 500 for ImageNet-1K and 217 for Places-434. For ImageNet-1K, the 500 class partition achieves 78.04\% top-1 (94.10\% top-5) accuracy and for Places-434, the 217 class partition achieves 55.06\% top-1 (84.05\% top-5) accuracy.

Using these datasets, we create three separate outlier detection problems that vary in difficulty:
\begin{enumerate}
    \item \textbf{Noise}: This represents the easiest problem, and has been commonly evaluated for the methods studied~\cite{hendrycks2017baseline, liang2018enhancing, krueger2018bayesian, lee2017training}. We generate synthetic images from a zero mean, unit variance Gaussian distribution to match the normalization scheme of training and test images.

    \item \textbf{Inter-Dataset}: As a problem of intermediate difficulty, we study each method's ability to detect outlier samples drawn from another large-scale dataset, i.e., for the ImageNet-1K trained method, samples are drawn from Places-434 and vice versa.  There are 18 classes that overlap between ImageNet-1K and Places-434.  These classes are removed from each OOD dataset. 

    \item \textbf{Intra-Dataset}: As the hardest task, the novel classes are made up of the remaining classes in each dataset.

    This is difficult because the image statistics of a class are often very similar to the statistics of other classes in the dataset; thus, minimizing the open space risk of any classification boundary is critical to achieve good bounded classification performance.
\end{enumerate}

\begin{figure*}[t]
    \centering
    \begin{subfigure}[b]{0.25\textwidth}
        \centering
        \includegraphics[width=\textwidth]{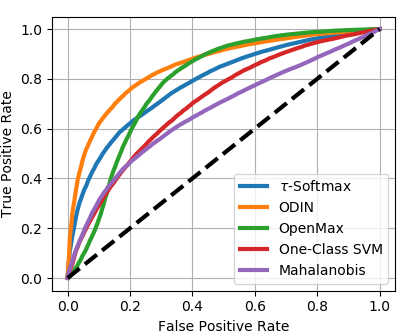}
        \caption{Cross-Entropy}
        \label{fig:auroc_ood_ce}
    \end{subfigure}
    \hspace{1mm}
    \begin{subfigure}[b]{0.25\textwidth}
        \centering
        \includegraphics[width=\textwidth]{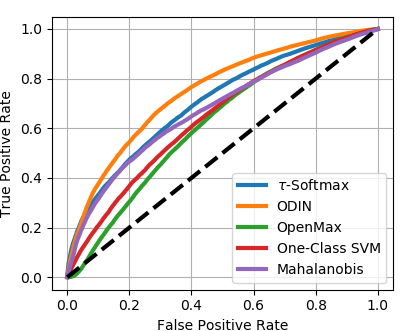}
        \caption{One-vs-Rest}
        \label{fig:auroc_ood_bce}
    \end{subfigure}
    \hspace{1mm}
    \begin{subfigure}[b]{0.25\textwidth}
        \centering
        \includegraphics[width=\textwidth]{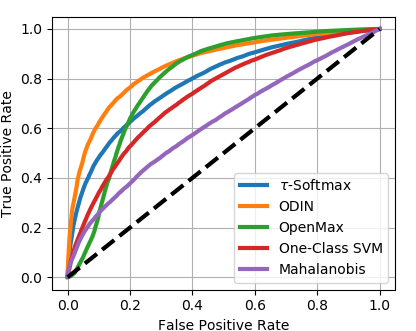}
        \caption{Background Class Reg.}
        \label{fig:auroc_ood_bkg}
    \end{subfigure}
    \caption{ROC curves for the \textbf{ImageNet / Intra-Dataset (Open Set)} test. See supplemental material for additional curves.}
    \label{fig:auroc_plots}
    \vspace{-1.5em}
\end{figure*}

The training set and models are kept fixed across the three paradigms, but the test sets vary across them. We construct the OOD evaluation sets for each problem/dataset by randomly choosing 10,000 in-distribution samples evenly among the in-distribution classes and 10,000 outlier samples evenly among the OOD classes within each respective dataset's validation set.

\subsection{Results}

We use the Area Under the ROC Curve (AUROC) metric to assess OOD detection performance of each approach as a binary detector for in and out samples. AUROC characterizes the performance across the full range of threshold values, regardless of the range of unique values for each inference method's scoring function. AUROC has been a commonly used metric for OOD detection in image classification datasets ~\cite{hendrycks2017baseline,liang2018enhancing,lee2017training,lee2018simple,hendrycks2018deep}. Our main results for each experimental paradigm and dataset are given in Table~\ref{tab:results_w_delong}. We then summarize the results for each experiment in Table~\ref{tab:summary-results} by computing the mean of each method's performance over various OOD tests and datasets. 

\subsubsection{Cross-Entropy Feature Space}
Overall, we see that ODIN performs best on detecting outliers to the ImageNet dataset, but the results on Places are much more varied. The performance of inference methods that directly apply thresholds to the output of the network appear to be strongly correlated with the overall closed set accuracy of the baseline model. For example, baseline methods, DOC, and ODIN all struggle to separate synthetic noise from in-distribution samples for the Places-434 dataset because the overall top-1 accuracy of the model is low.  Conversely, while the Mahalanobis method is able to accurately separate noise from the 217 classes in the Places experiment, it struggles to accurately separate noise from the 500 classes of ImageNet. 

\subsubsection{One-vs-Rest Feature Space}
ODIN performs best for all three experiments for ImageNet, OpenMax and Mahalanobis method perform best for Places. Thresholding methods performed well on noise experiments on ImageNet and while baseline methods struggled to separate noise from in-distribution samples on Places for the cross-entropy feature space, we see that it performed much better for the one-vs-rest space. Mahalanobis was again a top performer for inter and intra-dataset expermiments on Places. 

\subsubsection{Background Class Regularization Feature Space}
This feature space yielded most of the top performances over all experiments and feature spaces; however the level of benefit gained over standard cross-entropy training decreases as the OOD detection problem becomes more difficult (inter vs intra-dataset). ODIN was the top performer across all experiments in this feature space, but several other inference methods also performed well. The one-class SVM was a top performer on both inter and intra-dataset experiments indicating that separation between in-distribution and OOD samples in deep feature space was sufficient for a single non-linear, non-class conditional decision boundary to separate. 

\begin{table}
\caption{Mean summary statistics for each inference method across the three feature spaces tested.}
\label{tab:summary-results}
\centering
\resizebox{\linewidth}{!}{%
\begin{tabular}{lcccc}
\toprule
 & \textsc{Cross} & \textsc{One-} & \textsc{Bckgd.} &  \\
\textsc{Method} & \textsc{Entropy} & \textsc{vs-Rest} & \textsc{Reg.} & \textsc{Mean} \\
\midrule
$\tau$-Softmax & 0.715 & 0.704 & 0.844 & 0.754 \\
$\tau$-Sigmoid & 0.634 & \textbf{0.764} & 0.795 & 0.731 \\
ODIN & 0.719 & 0.729 & \textbf{0.901} & 0.783 \\
OpenMax & \textbf{0.804} & 0.718 & 0.835 & 0.731\\
One-Class SVM & 0.762 & 0.756 & 0.870 & \textbf{0.796} \\
Mahalanobis & 0.659 & 0.707 & 0.493 & 0.619 \\
\midrule
Mean & 0.715 & 0.702 & \textbf{0.790} & 0.736 \\
\bottomrule
\end{tabular}
}
\vspace{-1.5em}
\end{table}
\subsubsection{Additional Experiment: Model Depth and Width}
Current state-of-the-art networks on large-scale image datasets often have hundreds of layers and hundreds of convolutional filters per layer.  Previous work has shown that deeper and wider networks produce more accurate results but often lead to uncalibrated predictions~\cite{guo2017calibration}.  

We found that the relationship for OOD detection performance and model capacity more closely matches model accuracy and we do not observe a relative change in OOD detection performance for two models with different depths but similar closed-set accuracy.  We include results and a summary chart of performance versus model capacity in the supplemental material.

\section{Discussion and Conclusion}
Research in OOD detection has largely focused on either developing inference strategies for pre-trained models or a feature representation strategy for baseline inference methods for detecting OOD samples.  However, as our results show a large performance increase can be gained by combining an advanced inference technique with a feature space regularization strategy.  Nevertheless, the performance increase over baseline techniques appears to be much smaller as the dataset becomes more complex and the novelty detection problem becomes more difficult.  In Fig.~\ref{fig:auroc_plots} we show the resulting ROC curves for the ImageNet Intra-Dataset problem, which demonstrate that there is little to no benefit from background class regularization versus standard cross-entropy training in the open set recognition task.

In this paper, we performed a comprehensive comparison of outlier detection schemes for CNNs using large-scale image classification datasets. We organized methods into inference and feature space regularization strategies and outlined the general applicability of these methods.  Additionally, we established a testing paradigm with varying difficulty using different outlier datasets. Through this paradigm, we demonstrated that novelty detection performance is very dataset dependent but generally decreases as the similarity between the in-distribution and OOD classes decreases. Finally, there is still difficulty adapting current state-of-the-art feature representation strategies for large-scale datasets to work in accordance with advanced inference methods. Ultimately, challenges remain in adapting OOD detection methods for large-scale datasets and producing reliable recognition of novel inputs.

\ifthenelse{\boolean{ack}}{
\paragraph*{Acknowledgments.} This work was supported in part by DARPA/MTO Lifelong Learning Machines program [W911NF-18-2-0263]  and AFOSR grant [FA9550-18-1-0121]. We thank NVIDIA for the GPU donation. The views and conclusions contained herein are those of the authors and should not be interpreted as representing the official policies or endorsements of any sponsor. 
}{}
\small
\bibliographystyle{ieee}
\bibliography{library}
\ifthenelse{\boolean{combined}}{
\clearpage
\begin{center}
    {\Large Supplemental Material \normalsize}
\end{center}
\section*{S1. Additional Experiment: Model Capacity Impact on Performance}
As an additional experiment, we desired to understand if there was a correlation between model capacity in a CNN, i.e., the depth and width of convolutional layers, and the resulting OOD detection performance.  Our results in the main paper indicate that performance is related to overall model accuracy and varies as the feature space representation changes.  To answer this question we trained a series of ResNet models with either a fixed convolutional filter width (64) and varying depths (10-152 layers) or fixed depth (18 layers) and varying number of filter channels per layer (16-128).  This is the same protocol previously implemented for showing the disconnect between model capacity and confidence calibration~\cite{guo2017calibration}.  The results from these experiments are shown in Fig.~\ref{fig:model_capacity}.  As the results show, performance on detecting OOD images largely tracks overall model accuracy.  Thus as the depth and width grow and model accuracy increases, then outlier detection performance also increases.  Unlike the previously reported negative effect of model capacity on confidence calibration, there is no indication that increasing model depth or width negatively impacts OOD detection performance. 

\begin{figure*}[t]
    \centering
    \begin{subfigure}[b]{0.4\textwidth}
        \centering
        \includegraphics[width=\textwidth]{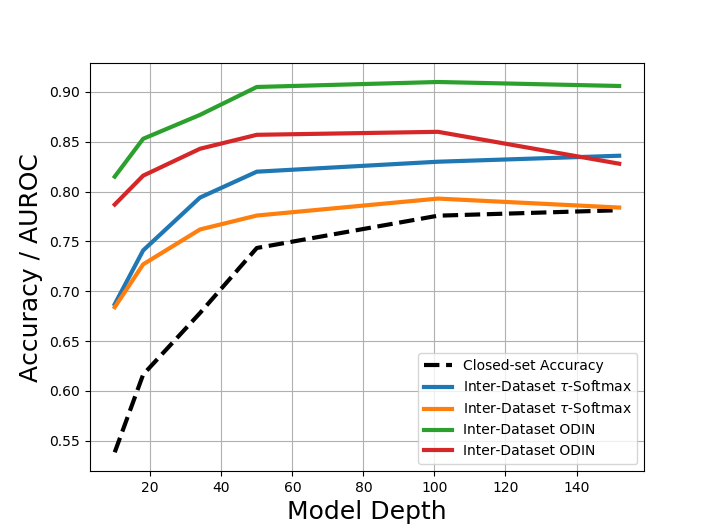}
        \label{fig:model_depth}
    \end{subfigure}
    \hspace{3mm}
    \begin{subfigure}[b]{0.4\textwidth}
        \centering
        \includegraphics[width=\textwidth]{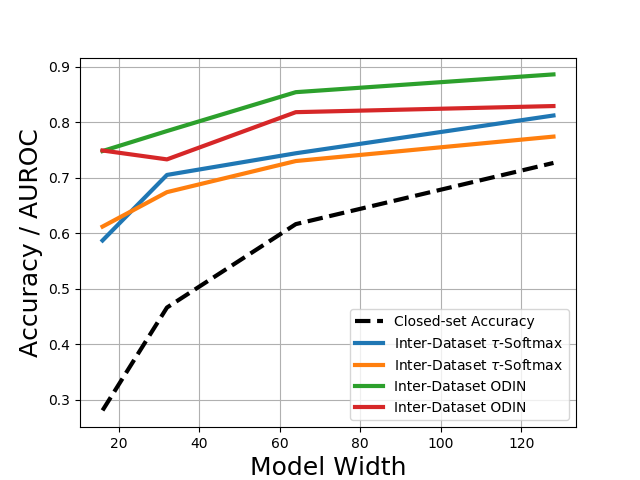}
        \label{fig:model_width}
    \end{subfigure}
    \caption{Examination of OOD detection performance as a function of model capacity.  A ResNet architecture was varied in either depth or width and trained on the ImageNet-500 split and then tested for detecting image classes unseen during training via either the Places dataset (Inter-dataset) or the remaining ImageNet categories (Intra-dataset). Overall improvements in performance as reflected in the AUROC of the model track improvements in model accuracy as model capacity increases.}
    \label{fig:model_capacity}
    \vspace{-1.5em}
\end{figure*}

\section*{S2. Additional Metrics for Empirical Results}

While the AUROC metric is a good high level measure of performance for OOD detection, we also desire to understand how performance on accepting or rejecting samples affects the overall classification performance.  To address this, we also adopt the Open Set Classification (OSC) metric~\cite{dhamija2018reducing}, which is an adaptation on the traditional ROC curve that plots the correct classification rate versus false positive rate.  This correct classification rate is the difference between the model accuracy and the false negative rate. Intuitively, this metric takes into account whether true positive samples are actually classified as the correct class and thus rewards methods which reject incorrectly classified positive samples before rejecting samples that are correctly classified.  We calculate the area under the OSC curve (AUOSC) to provide an easy assessment of performance across different experimental paradigms and datasets.

Comparing to AUROC, which gives the likelihood that an in-distribution sample will have a higher acceptance score than an OOD sample, the AUOSC shows the reduction in closed-set classification performance from implementing the additional step in OOD detection of determining whether a sample is in or out.  We further adapt the AUOSC metric by dividing by the closed-set classification accuracy to give a normalized AUOSC value that is the percentage of closed set classification performance (on average) achievable if the OOD detection technique is implemented.  We believe that these additional metrics give a network designer additional insight into what the costs of implementing a OOD detection method to a traditional classification model.

\begin{table*}[t]
\centering
\caption{AUOSC and Normalized-AUOSC results averaged over 5 runs for the methods tested.  Top three performers for each in-distribution / out-of-distribution combination are in progressive shades of \colorbox[HTML]{a4c2f4}{blue}.}
\label{tab:results_auosc}
\resizebox{\textwidth}{!}{%
\begin{tabular}{c|r|ccc|ccc|}
\multicolumn{1}{l|}{} & \multicolumn{1}{l|}{} & \multicolumn{3}{c|}{\textbf{ImageNet}} & \multicolumn{3}{c|}{\textbf{Places}} \\ \hline
\textbf{Features Space} & \multicolumn{1}{l|}{\textbf{Inference Method}} & \textbf{\begin{tabular}[c]{@{}c@{}}Gaussian \\ Noise\end{tabular}} & \textbf{\begin{tabular}[c]{@{}c@{}}Inter-Dataset \\ (OOD)\end{tabular}} & \textbf{\begin{tabular}[c]{@{}c@{}}Intra-Dataset \\ (Open Set)\end{tabular}} & \textbf{\begin{tabular}[c]{@{}c@{}}Gaussian\\ Noise\end{tabular}} & \textbf{\begin{tabular}[c]{@{}c@{}}Inter-Dataset\\ (OOD)\end{tabular}} & \textbf{\begin{tabular}[c]{@{}c@{}}Intra-Dataset\\ (Open Set)\end{tabular}} \\ \hline
\multirow{6}{*}{Cross-Entropy} 
 & $\tau$-Softmax & 0.750/0.995 & 0.673/0.893 & 0.652/0.864 & 0.353/0.671 & 0.385/0.733 & 0.368/0.700 \\
 & DOC & 0.753/0.999 & 0.676/0.897 & 0.587/0.779 & 0.207/0.394 & 0.397/0.754 & 0.400/0.762 \\
 & ODIN & \td 0.754/1.000 & 0.710/0.941 & \sd 0.676/0.897 & 0.208/0.395 & 0.399/0.759 & \td 0.405/0.771 \\
 & OpenMax & 0.753/0.999 & 0.159/0.211 & 0.071/0.094 & 0.460/0.875 & 0.316/0.601 & 0.256/0.487 \\
 & One-Class SVM & 0.744/0.987 & 0.632/0.838 & 0.537/0.713 & 0.449/0.853 & 0.355/0.676 & 0.349/0.665 \\
 & Mahalanobis & 0.675/0.896 & 0.461/0.611 & 0.526/0.697 & \td 0.526/1.000 & 0.356/0.678 & \ft 0.439/0.751 \\ \hline
\multirow{6}{*}{One-vs-Rest} 
 & $\tau$-Sigmoid & 0.649/0.999 & 0.539/0.830 & 0.521/0.801 & 0.410/0.814 & 0.374/0.741 & 0.358/0.709 \\
 & DOC & 0.633/0.974 & 0.483/0.744 & 0.470/0.724 & 0.505/1.000 & 0.389/0.772 & 0.381/0.755 \\
 & ODIN & 0.650/1.000 & 0.560/0.862 & 0.518/0.797 & 0.348/0.690 & 0.389/0.771 & 0.392/0.777 \\
 & OpenMax & 0.649/0.999 & 0.500/0.769 & 0.474/0.729 & 0.489/0.968 & 0.375/0.746 & 0.375/0.728 \\
 & One-Class SVM & 0.637/0.981 & 0.499/0.768 & 0.418/0.643 & 0.438/0.868 & 0.362/0.717 & 0.361/0.716 \\
 & Mahalanobis & 0.623/0.959 & 0.439/0.676 & 0.463/0.712 & 0.505/1.000 & 0.316/0.626 & 0.338/0.670 \\ \hline
\multirow{6}{*}{Background Class Regularization} 
 & $\tau$-Softmax & 0.730/0.965 & 0.717/0.948 & \td 0.659/0.871 & \ft 0.557/0.998 & \sd 0.510/0.914 & 0.403/0.721 \\
 & DOC & \sd 0.755/0.999 & \ft 0.740/0.978 & \td 0.659/0.872 & \ft 0.557/0.998 & \sd 0.510/0.914 &  0.402/0.721 \\
 & ODIN & \ft 0.756/1.000 & \sd 0.739/0.977 & \ft 0.682/0.901 & \ft 0.558/1.000 & \ft 0.528/0.945 & \sd 0.413/0.739 \\
 & OpenMax & \sd 0.755/0.999 & 0.672/0.888 & 0.279/0.368 & 0.553/0.990 & 0.506/0.907 & 0.399/0.715 \\
 & One-Class SVM & 0.743/0.982 & \td 0.719/0.951 & 0.569/0.752 & \sd 0.531/0.988 & \td 0.508/0.945 & 0.385/0.715 \\
 & Mahalanobis & 0.687/0.909 & 0.280/0.371 & 0.488/0.645 & 0.442/0.822 & 0.119/0.222 & 0.339/0.630 \\ \hline
\end{tabular}
}
\end{table*}

\pagebreak

}

\end{document}